\documentclass[letterpaper, 10 pt, journal, twoside]{IEEEtran}

\IEEEoverridecommandlockouts
\usepackage{cite}
\usepackage{amsmath,amssymb,amsfonts}
\usepackage{caption} 
\usepackage{amssymb}
\usepackage{tikz}    
\usepackage{subcaption} 

\usepackage[normalem]{ulem}
\usepackage{amssymb}
\usepackage{pifont}
\usepackage[table,xcdraw]{xcolor}
\usepackage{algorithmic}
\usepackage{graphicx}
\usepackage{soul}
\usepackage{siunitx}
\usepackage{tabularx}
\usepackage{textcomp}
\usepackage{xcolor}
\usepackage{booktabs}
\usepackage[hyphens]{url}
\usepackage[colorlinks]{hyperref}

\newcommand{\hour}[1]{%
    \begin{tikzpicture}
    \draw (0,0) -- (6pt,0) -- (0,6pt) -- (6pt,6pt) -- (0,0);
    \end{tikzpicture}
}

\newcommand{\loz}[1]{%
    \begin{tikzpicture} 
    \draw (0,3.5pt) -- (3pt,7pt) -- (6pt,3.5pt);
    \draw (0,3.5pt) -- (3pt,0) -- (6pt,3.5pt);
    \end{tikzpicture}
}

\newcommand{\quadrato}[1]{%
    \begin{tikzpicture} 
    \draw (0,0) -- (0,6pt);
    \draw (0,6pt) -- (6pt,6pt);
    \draw (6pt,0) -- (6pt,6pt);
    \draw (6pt,0) -- (0,0);
    \end{tikzpicture}
}

\def\BibTeX{{\rm B\kern-.05em{\sc i\kern-.025em b}\kern-.08em
    T\kern-.1667em\lower.7ex\hbox{E}\kern-.125emX}}
\begin{document}

\title{A Distributed Multi-Modal Sensing Approach for Human Activity Recognition in Real-Time Human-Robot Collaboration\\

\thanks{This is the accepted version of a paper accepted to IEEE Robotics and Automation Letters (RA-L), to be presented at ICRA 2026.}
\thanks{
\scriptsize{$*$ Corresponding author \tt\footnotesize {valerio.belcamino@edu.unige.it}}}
\thanks{\scriptsize{$^{1}$ TheEngineRoom, Department of Informatics Bioengineering, Robotics and System Engineering, University of Genoa, Genoa, Italy.}}
\thanks{\scriptsize{$^{2}$ Faculty of Mechanical Engineering, The University of Danang–University of Science and Technology, Danang, 54 Nguyen Luong Bang, 550000,Da Nang, Vietnam }}
\thanks{\scriptsize{$^{3}$ Soft Haptics Lab, School of Materials Science, Japan Advanced Institute of Science and Technology, Nomi
923-1292, Japan }}%

\thanks{Digital Object Identifier (DOI): see top of this page.}

}
\DeclareRobustCommand{\IEEEauthorrefmark}[1]{\smash{\textsuperscript{\scriptsize #1}}}

\author{
Valerio Belcamino$^{*,1}$, Nhat Minh Dinh Le$^{2}$, Quan Khanh Luu$^{3}$, Alessandro Carfì$^{1}$, \\Van Anh Ho$^{3}$, Fulvio Mastrogiovanni$^{1}$}


\maketitle

\begin{abstract}
Human activity recognition (HAR) is fundamental in human-robot collaboration (HRC), enabling robots to respond to and dynamically adapt to human intentions. 
This paper introduces a HAR system combining a modular data glove equipped with Inertial Measurement Units and a vision-based tactile sensor to capture hand activities in contact with a robot.
We tested our activity recognition approach under different conditions, including offline classification of segmented sequences, real-time classification under static conditions, and a realistic HRC scenario.
The experimental results show a high accuracy for all the tasks, suggesting that multiple collaborative settings could benefit from this multi-modal approach.

\end{abstract}

\section{Introduction}\label{sec:intro}
One of the research objectives in Human-Robot Collaboration (HRC) is to achieve fluency comparable to human-human interaction \cite{dillman}. A fundamental component in pursuing this goal lies in the robot's ability to recognize user actions, i.e., human activity recognition (HAR), consequently acting appropriately \cite{1570204}.
Human activities encompass a wide range of interactions, some characterized by motions \cite{sl}, others by the application of contact forces \cite{grasp}, and many requiring the integration of both, where motions are guided by, or result, in applied forces \cite{1570207}.
For these reasons, a comprehensive HRC framework should include a human activity recognition system capable of detecting both movements and contact forces.

\begin{figure}[!ht]
\centering
\includegraphics[width=0.8\columnwidth, trim={0cm 4cm 0 0cm},clip]{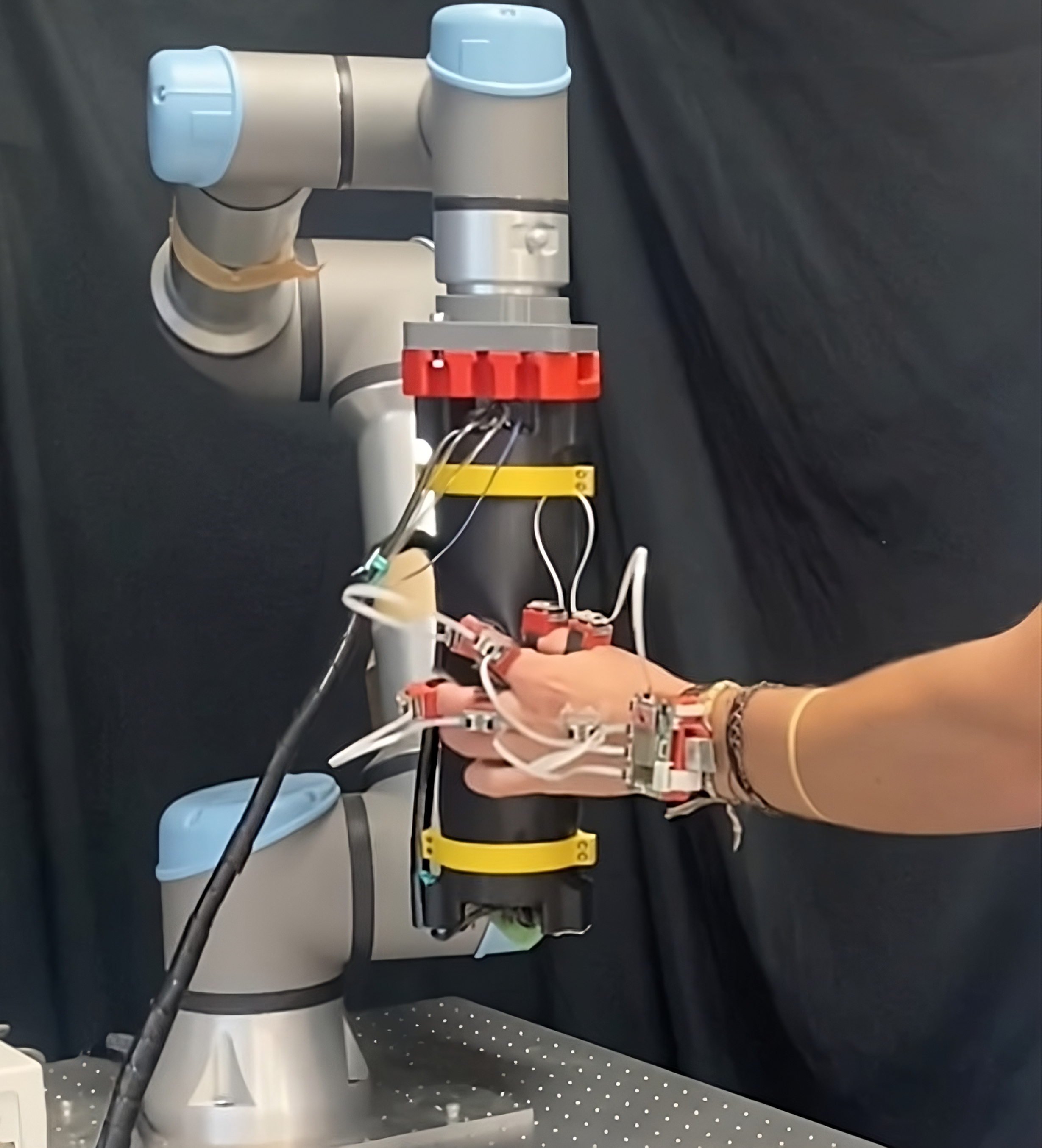}
\caption{The user pinches the TacLINK connected to the UR5 robot while wearing the TER glove.
}
\label{fig:frontpage}
\end{figure}

For human motion sensing, the literature proposes several approaches grouped into wearable \cite{wearables} and external \cite{external} devices.
External devices generally rely on optical sensors, including different technologies, from low-cost RGB cameras to professional motion capture systems. These systems are frequently used in HAR to track human joint states, which are then processed to identify activities \cite{bodytracking}. Alternatively, captured images can be directly analyzed to classify actions or processed with motion extraction algorithms, such as optical flow \cite{opticalflow}. Additionally, external optical devices support multi-person tracking \cite{multi} and can extract additional information on the environment through scene segmentation \cite{sceneseg}. On the other hand, these sensors have several limitations, including occlusion and sensitivity to changing or adverse lighting conditions.
Concerning wearables, the two most popular technologies \cite{review} are flex sensors \cite{flex} and inertial measurement units (IMUs) \cite{imu}. 
Flex sensors are usually sewn into suits \cite{flexSuit} or gloves \cite{flexGlove} and used to estimate the joint angles of the body from their mechanical deformation. They are a valid solution for joints with a single degree of freedom but generally struggle with the multi-axial ones. Additionally, they suffer from permanent deformation over time \cite{deformation}. Instead, IMUs estimate rigid body orientations and are often positioned in pairs on consecutive links to track the motion of a joint using their relative orientations \cite{IMUdouble}. They can track joints with multiple DOF \cite{shoulder} but suffer from drifting problems \cite{tokyodrift} and are affected by misalignments between the sensor and link orientations \cite{sensor2segment}.

\begin{table}[h!]
\caption{List of the actions used in our study; for each action, the columns show the Name, Contact Area (S/M/L), Pressure Intensity (L/M/H), Frequency (S/M/L), and a Description. For the idle action, contact area, pressure, and frequency are omitted since no contact occurs.}
\label{tab:action_characteristics}
\centering
\resizebox{\columnwidth}{!}{
\begin{tabular}{@{}lllll@{}}
\textbf{Action} & \textbf{CA} & \textbf{PI} & \textbf{F} & \textbf{Description} \\
\hline
Pinching    & S & H & M & Two fingers press a single point. \\
Pulling     & L & H & L & Gripping and slightly tugging. \\
Pushing     & L & H & L & Steady force with hand/fingers. \\
Rubbing     & M & M & H & Fingers move back and forth. \\
Patting     & L & M & H & Intermittent open palm contact. \\
Tapping     & M & M & H & Quick fingertip contacts. \\
Scratching  & S & L & H & Dragging fingers lightly. \\
Lingering   & M & L & L & Steady contact, no movement. \\
Massaging   & L & M & M & Circular pressure with palm. \\
Squeezing   & L & H & L & Compression using fingers. \\
Trembling   & S & L & M & Slight fingertip oscillations. \\
Shaking     & L & M & H & Rapid large oscillations. \\
Stroking    & M & L & M & Smooth palm motion. \\
Poking      & S & M & M & Localized finger force. \\
Idle        & – & – & – & No physical interaction. \\
\end{tabular}}
\end{table}

As for perceiving contact forces, the development and study of tactile sensing have been valuable in HAR, especially for activities involving physical interaction with objects or surfaces. 
Commonly, tactile sensors rely on resistive, capacitive, and piezoelectric technologies \cite{pressurereview}, producing varying intensities of current based on the applied pressure. These solutions, such as flex sensors, are easy to embed in clothing in areas frequently involved in physical contact \cite{review}. For example, in gloves, tactile sensors are commonly positioned on fingertips and palms \cite{pressureGlove}. In HRC, research in tactile sensing led to the development of robotic skin \cite{skin}, which enables robots to detect and respond to physical interactions with humans \cite{handhuman}. Additionally, they play a crucial role in human-robot handover tasks, where tactile feedback enables robots to securely and intuitively exchange objects with other humans or robots \cite{1570204}. Another approach to tactile sensing involves vision-based tactile sensors, which provide distributed tactile feedback by capturing images of deformations in a soft membrane.
Like traditional tactile sensors, they can estimate force and contact information \cite{softtactilerev}. Furthermore, their soft, deformable design also enhances safety in HRC by attenuating contact forces and reducing the risk of injury during close interactions \cite{quanControl}. 

The literature on HAR includes various multimodal approaches that aim to leverage motion and contact force perception \cite{wang2024multimodal}. Multimodal perception of human activities has been applied in many fields and for different tasks, such as gait analysis \cite{s140203362}, studying human interaction with musical instruments \cite{piano},
supporting gestural interfaces in human-computer interaction \cite{hci}, and early detection of motion impairments \cite{impairmentMixed}. In robotic applications, extensive work has been done primarily on handover detection \cite{handoverReview}, utilizing RGB cameras or motion capture systems for hand tracking, alongside tactile or torque sensors for detecting contact forces \cite{thoduka2024multimodalhandoverfailuredetection,10.1145/3656650.3656675}. \cite{thoduka2024multimodalhandoverfailuredetection, handoverMocap, 10.1145/3656650.3656675}. 
Nevertheless, outside the specific context of handover tasks, the integration of contact forces and human motion for recognizing human activities and improving human-robot interactions remains largely unexplored.

Here, we focus on HAR for collaborative robotics, merging motion and contact force perception in a classification system for hand activities. We decided to employ IMUs for motion tracking due to their ability to track multiple degrees of freedom, which is crucial for capturing the complex hand kinematics, and their resilience to occlusions, essential for tasks involving contact. As for tactile perception, we rely on a vision-based device, which acts as an additional sensorized link for the robot. In particular, our system combines a vision-based tactile sensor equipped on the robot\cite{zhang2022hardware} and an IMU-based data glove. By combining these two sensory inputs, our system enables discrimination among a wide range of hand activities with distinct motion and contact patterns, enhancing tactile perception through the complementary hand-tracking capabilities offered by the glove.
We benchmarked our method by classifying 15 actions commonly used in tactile sensing research \cite{touchactions2} and social interaction \cite{socialtouch}, covering a range of contact characteristics such as contact surface, pressure and frequency. After validating our classifier on a segmented dataset, we transitioned to an online classification. Finally, we deployed our solution in an HRC scenario, where the TacLINK was mechanically attached to a UR5 robot. The developed HAR solution mediated the interaction with the human by adjusting the robot's trajectory in response to the user's hand activities, demonstrating the system's potential for safe and responsive collaboration.

\section{Methodology}
\label{sec:methodology}
Our human activity recognition system leverages sensors distributed between the human and the robot. We utilized the TacLINK \cite{taclink1}, a large-scale vision-based tactile sensor constructed from silicon polymer,  for contact force perception. The device has a cylindrical shape with a height of \SI{23}{\cm} and a radius of \SI{4}{\cm}, providing a total active sensing area of \SI{578.05}{\cm^2}. The inner surface of the sensor is covered in equally spaced arrays of reflective markers, and two RGB cameras are positioned at the top and bottom of the cylinder. These cameras capture the position of markers to track deformations on the device's surface, enabling the estimation of both the contact area and the magnitude of contact forces. The TacLINK outputs two RGB video streams at a fixed framerate of 30 fps with a resolution of 1080p. This device can be attached to a robot manipulator as an additional link, as illustrated in Fig. \ref{fig:frontpage}. We also adopted the TER glove \cite{terglove}, a data glove equipped with 9-axis IMUs for sensing human motions. The device's modular architecture allows for adjusting the number of IMUs, up to 12 sensors, based on application needs. However, increasing the number of modules affects sampling frequency and battery life. Given the frame rate of the TacLINK and considering that our set of actions presents a significant coupling between the middle, ring and little fingers, we decided to adopt a configuration with eight sensors ensuring a frequency of \SI{40}{\hertz} and about one hour of autonomy. As shown in Fig. \ref{fig:frontpage}, the thumb, index, and middle fingers each were equipped with two sensors, while the remaining sensors track the hand back and the wrist.
At runtime, the glove streams the three-axis accelerations, angular velocities and magnetic field values for each of the eight modules to our workstation via Wi-Fi.

\begin{figure}[t]
\centering
\includegraphics[width=0.8\columnwidth]{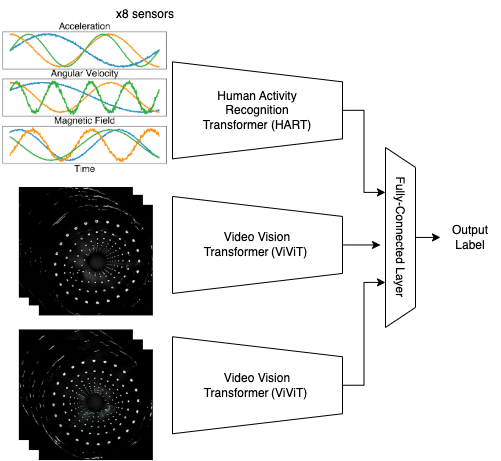}
\caption{The adopted neural network architecture is composed of three separate branches merged in the last layer. The top branch, based on HART, takes as input the raw data from the IMUs, and the other two, based on the ViViT, work on the video streams from the TacLINK.}
\label{fig:model}
\end{figure}

To fuse the two sensor sources, we first downsampled the sampling rate of the IMUs to match those of the cameras, aligning each IMU reading with a corresponding camera frame using synchronized timestamps. Then, we processed the data with a custom neural network architecture leveraging a late fusion approach \cite{latefusion}. A late fusion approach was chosen over early fusion due to its flexibility in processing heterogeneous data types independently through specialized branches, allowing each data modality to contribute with high-level features before merging \cite{latefusion2}. This approach enables the network to focus on distinct motion patterns from IMUs and tactile features from camera images by processing each modality independently until the final decision stage. By isolating their contributions, the model optimizes spatial-temporal patterns unique to each sensor type, enhancing accuracy in classifying complex actions involving motion and contact.

Our architecture builds on the Video Vision Transformer (ViViT)\cite{arnab2021vivitvideovisiontransformer} and Human Activity Recognition Transformer (HART)\cite{s22051911} architectures, combining them in a multi-branch framework. As shown in Fig. \ref{fig:model}, the architecture consists of three parallel branches: two ViVit backbones for each video input and one HART for the IMU data.
Each ViViT branch processes one of the two video streams from the TacLINK, focusing on capturing spatial-temporal patterns in tactile data. ViViT extracts detailed representations of the hand's interaction with the sensor surface, identifying specific contact shapes, pressures, and deformations. The third branch, based on the HART structure, processes the IMU data from the TER glove. HART is well-suited for temporal sequences, capturing sequential patterns in hand and joint movements. This branch allows the model to identify motion-related features related to heading, velocity profiles and inter-joint synergies, essential for identifying dynamic activities.
In the late fusion step, the features extracted from all three branches are stacked and fed into a fully connected layer. This final layer integrates information from all modalities, performing the classification based on a comprehensive understanding of contact forces and motion cues.

\section{Experimental Setup}
\label{Experimental Setup}
To evaluate our approach, we focused on classifying 15 distinct actions, selected to encompass a broad range of contact forces and motion characteristics, including variations in contact area, intensity, and frequency of interaction. These hand activities are representative of common gestures in everyday life and human–robot or human–human interactions; as such, their reliable recognition is crucial for natural and effective collaboration. The list of actions includes \textit{pinching, pulling, pushing, rubbing, patting, tapping, scratching, lingering, massaging, squeezing, trembling, shaking, stroking, poking and idle}. Table \ref{tab:action_characteristics} shows, for each action, a brief description and an overview of their contact properties. Furthermore, a video with examples for each action is available online\footnote{\footnotesize{\href{https://www.youtube.com/watch?v=RmqrACKARa8}{youtube.com/watch?v=RmqrACKARa8}}}.
We evaluated our approach across three distinct testing scenarios: offline, online and dynamic validation.

\begin{figure*}[!t]
\centering
\includegraphics[width=0.75\textwidth]{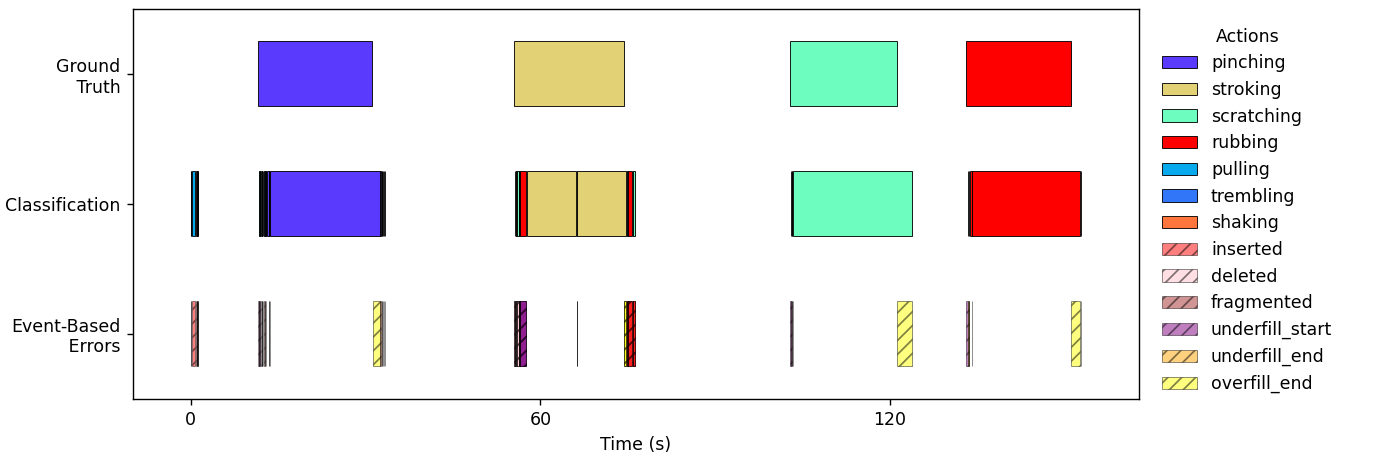}
\caption{The classification of a recording segment from the continuous dataset. The first row defines the ground truth labels, while the second one depicts the output of the classifier. The last row shows the event-based error metrics described in Section \ref{Experimental Setup}. For visual clarity, the idle action is not associated to a color and it is represented by the white spaces between the blocks.}
\label{fig:timeplot}
\end{figure*}

\subsection{Offline Validation}
We designed the first scenario to collect training data and preliminary evaluate offline classification of pre-segmented actions. In this case, the TacLINK was vertically mounted on a table, and we recorded examples for each action separately, creating a segmented dataset of the 15 classes. For each action, participants were first shown the gesture and then asked to repeat it several times during a 1-minute recording using the hand on which they wore the glove. 
We then repeated the experiment, moving the glove to the opposite hand. Participants could touch the TacLINK at any point on its sensorized surface. This process resulted in a dataset of 30 minutes per subject, with 8 participants, resulting in approximately 4 hours of recorded data.

Given the TacLINK's symmetrical, circular shape, we cropped all images to a squared region centered on the outer ring of markers, removing unnecessary borders from the 16:9 aspect ratio. The cutting mask is derived from the TacLINK's geometry and applied uniformly across all tests, ensuring that no contact occurs outside the masked region. The result of this processing step is displayed by the frames input to the ViViT branches in Fig. \ref{fig:model}.
Additionally, we converted the colour channels to greyscale and resized the resolution to 224x224 pixels, chosen as a trade-off between the increased training performance and the reduced image quality. Furthermore, we divided each 1-minute recording into 3-second sequences, resulting in over 4,000 samples. We chose a 3-second duration because it is long enough to capture even the lengthiest actions. However, since the segmentation was automatic, there could be sequences where actions are cut off or, in some cases, no action is contained within the sequence if the cut coincides with idle time between two consecutive executions of the same action. Finally, we standardized the video frames by calculating the mean and standard deviation for each split and normalized the IMU data using the full-scale range of the sensors. We divided the resulting dataset into three partitions following these proportions: 60\% for the training set, 20\% for validation and 20\% for testing.

\subsection{Online Validation}
The second test focused on the capability of our model to perform classification online using continuous sequences of human interactions. In an experimental setup identical to the previous one, we collected a second dataset of multiple action sequences. Specifically, we employed 4 participants, who performed all 15 actions in a continuous sequence interspersed with small pauses. We established a priori the order in which volunteers performed the actions to ensure a balanced dataset. Each volunteer performed the sequence twice, once with each hand, resulting in trials of approximately 15 minutes each and a total dataset length of around 2 hours. Two participants from this experiment were previously present in the first one.
An experimenter manually annotated the resulting continuous sequences to ensure accurate labelling. The annotation process utilizes the predetermined order of actions and the video streams from TacLINK's cameras, allowing the start and end of each action to be easily identified.
Intervals between actions were assigned to the \textit{idle} class.
We fed each sequence of this dataset into our neural network, previously trained on the segmented data, and we collected the output labels for each frame. 
Specifically, we used a moving window approach, where a buffer is continuously updated with new sensor samples. We chose a window of 90 samples to match the 3-second segments of the offline dataset.
The same processing transformations described for the first dataset have been applied in this case at each timestep of the sequence, including cropping, scaling, greyscale conversion, and standardization for the images and normalization with the full-scale ranges for the IMU data.

To evaluate the model performance in online classification, we compared its output labels to the ground truth. We began with a naive frame-by-frame approach and subsequently employed more detailed event-based metrics proposed by J.Ward et al., 2010 \cite{ward}. These metrics are based on events, i.e., complete instances of an action, delimited by a start and end time. 
Compared to frame-by-frame analysis, event-based analysis is preferable as it identifies errors while classifying them into different categories based on their position within the sequence. Possible event-based errors are \textit{insertion, merge, overfill, deletion, fragmentation} and \textit{underfill}. \textit{Insertion} errors occur when an extra action is detected where none exists. \textit{Merge} errors happen when two or more distinct actions are combined. \textit{Overfill} errors indicate that the detected action extends beyond the actual event’s end time, while \textit{deletion} errors occur when an action goes entirely unrecognized. \textit{Fragmentation} errors divide a single action into multiple detections, and \textit{underfill} errors occur when the detected action covers less than the actual event duration. Additionally, \textit{overfill} and \textit{underfill} can be further specialized to describe errors occurring at the start or end of a specific event (e.g., \textit{underfill\_start} to address missing frames at the beginning of an action and \textit{underfill\_end} for missing frames at its end). Lastly, the event-based analysis also considers \textit{true positives} and \textit{true negatives}. 
An example of the usage of these metrics can be observed in Fig. \ref{fig:timeplot}, which demonstrates different kinds of event-based errors.
An additional benefit of these metrics is that, depending on the task, not all error classes have the same importance. For instance, in a collaborative robotics setting where a robot must respond promptly to specific activities, deletion errors (missing an action) may be critical, as they could lead to inaction when a response is needed. 
In contrast, fragmentation errors, where a single action is divided into multiple detections, are generally less disruptive since they do not hinder the recognition of the action itself, although they may slightly impact reaction timing.

\begin{figure}[t]
    \centering
    \begin{subfigure}[b]{0.38\linewidth}
        \centering
        \includegraphics[width=\linewidth]{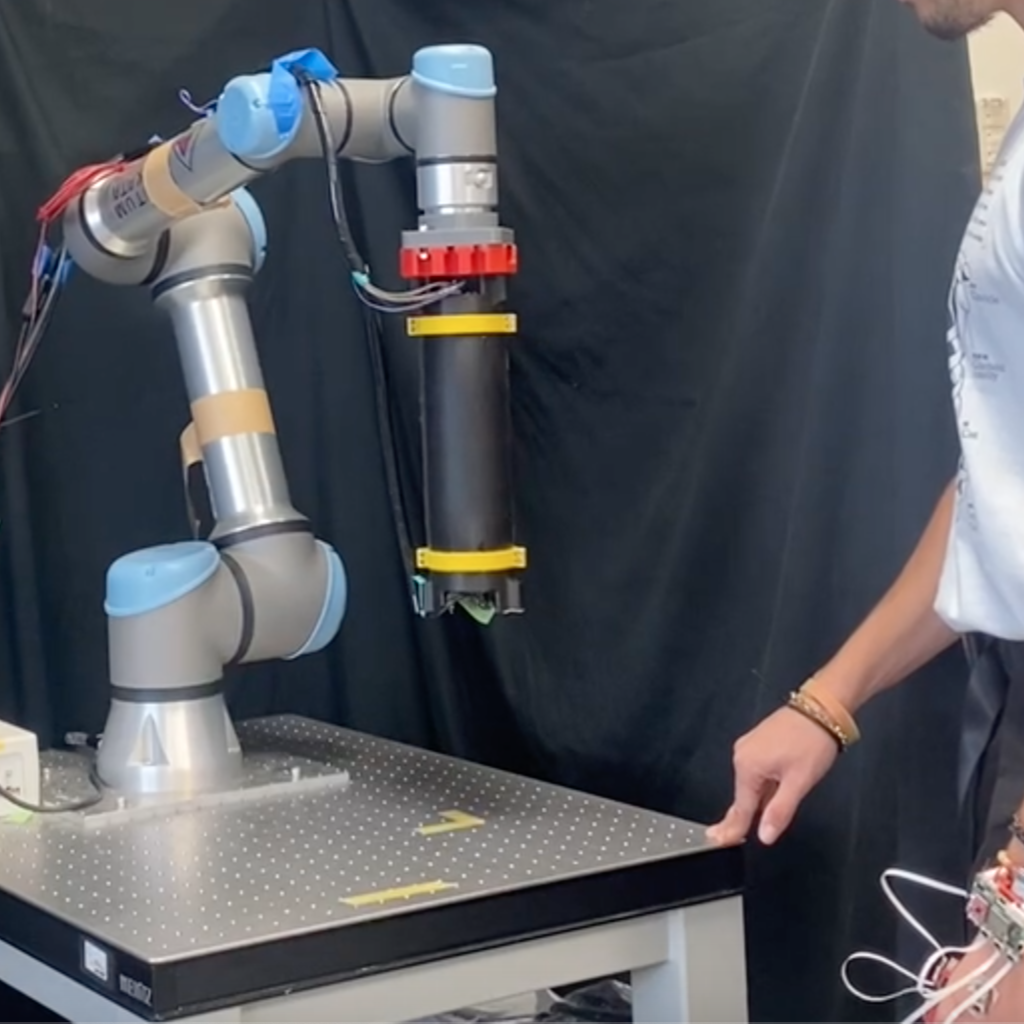}
        \label{fig:image1}
    \end{subfigure}
    \hspace{0.01cm}
    \begin{subfigure}[b]{0.38\linewidth}
        \centering
        \includegraphics[width=\linewidth]{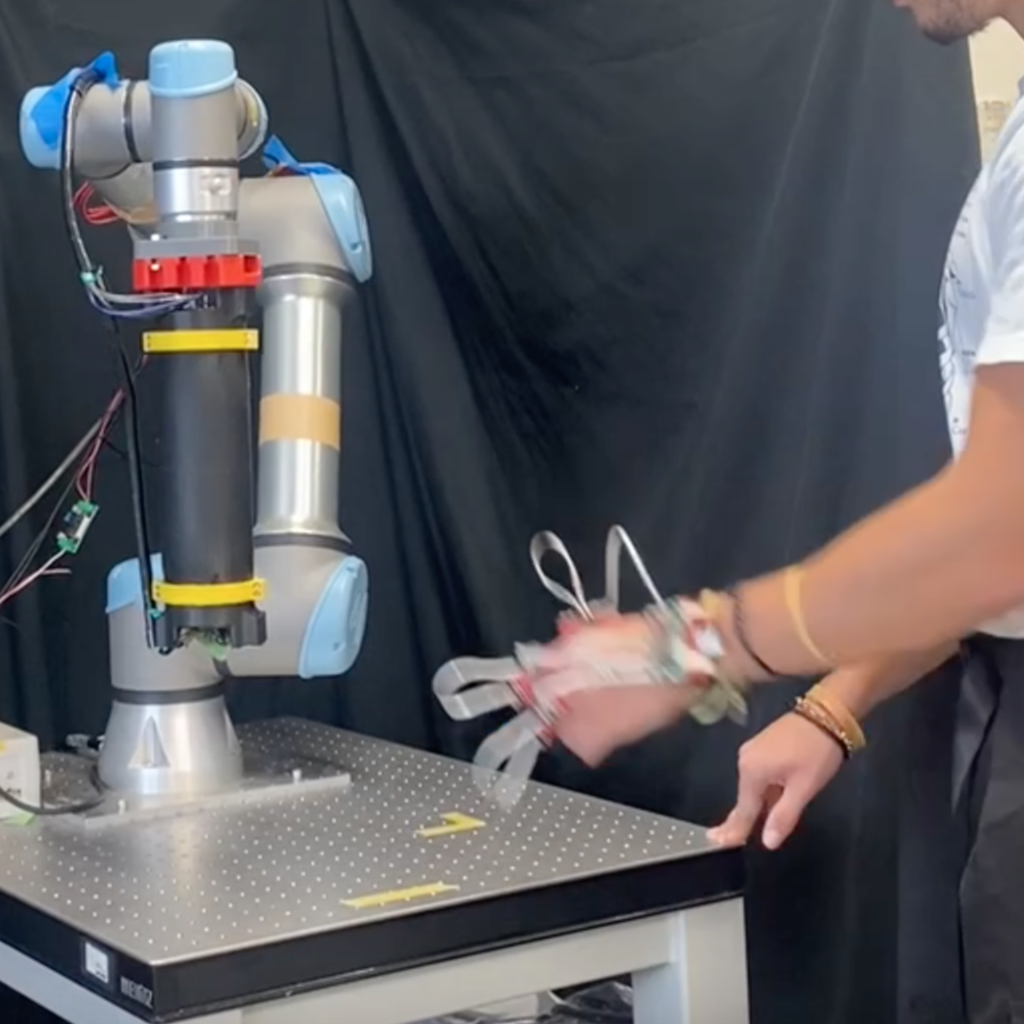}
        \label{fig:image2}
    \end{subfigure}
      \par\vspace{-0.25cm}
    \begin{subfigure}[b]{0.38\linewidth}
        \centering
        \includegraphics[width=\linewidth]{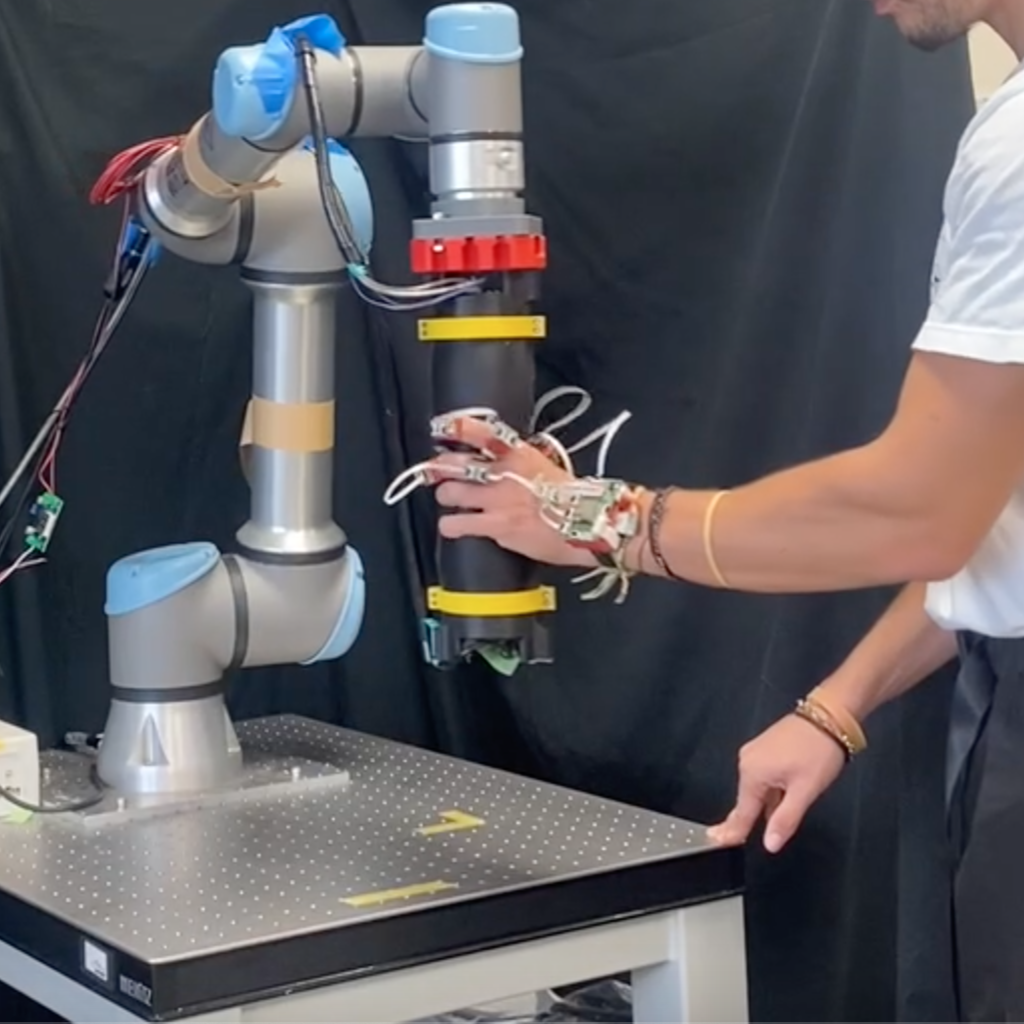}
        \label{fig:image3}
    \end{subfigure}
    \hspace{0.01cm}
    \begin{subfigure}[b]{0.38\linewidth}
        \centering
        \includegraphics[width=\linewidth]{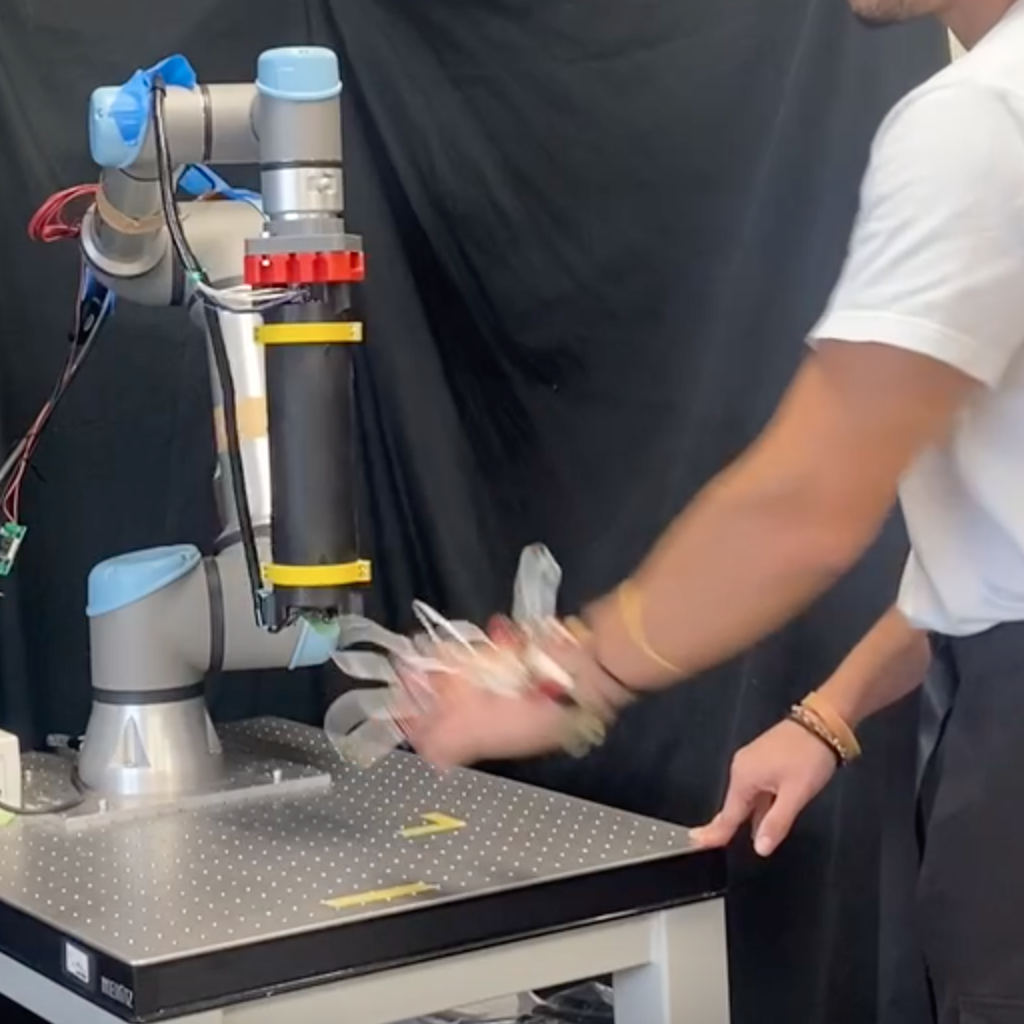}
        \label{fig:image4}
    \end{subfigure}

    \caption{Snapshots from a Dynamic Validation trial. The user waits while the robot follows its trajectory (top left), receives instructions and approaches the robot (top right), performs the required action on the TacLINK (bottom left), and returns to rest after the model recognises the action (bottom right).}
    \label{fig:2x2grid}
\end{figure}

\subsection{Dynamic Validation}

In the last experimental scenario, we mounted the TacLINK sensor on a UR5 robotic arm (see Fig. \ref{fig:frontpage}) to test the ability of our model to work in dynamic conditions. To simulate a realistic HRC use case, we designed five predefined robot trajectories, each representing a different geometric shape traced on a horizontal plane by the UR5. The five trajectories that we selected are: square ( $\square$ ), diamond ( \loz{}), hourglass ( \hour{}), triangle ( $\triangle$ ) and circle ( $\bigcirc$ ). This set of shapes encompasses different velocity profiles and distinct patterns of directional changes intended to simulate general motions that can occur during human-robot collaborations. Additionally, we programmed these movements to cover the entire workspace, forcing the user to follow the link, extending their arms or leaning forward. 
These trajectories were pre-recorded to avoid delays from real-time planning and to eliminate potential planning inconsistencies. Although the trajectory shapes were fixed, their execution order was randomly selected at runtime to preserve variability in the experiment. We limited the robot’s motion to a 2D plane, both to align with the tabletop setup used for dataset acquisition and to reflect common HRI scenarios such as pick-and-place tasks.

Before the experiment, each participant was shown the full set of possible actions and informed that their task would be to perform five of them—each intended to influence the robot’s trajectory.
During the experiment, the robot operated in a closed-loop manner, continuously following predefined trajectories while monitoring the human’s actions. Upon recognizing the execution of a specified action, it responded by transitioning to the next trajectory.

Each trial concluded after five transitions, with action-trajectory pairings randomized for balance across instances. 
The experiment involved 10 participants, half using their right 
hand, four of whom also contributed to the first dataset. We recorded the time elapsed between the experimenter instruction and the robot transition to a new trajectory as a proxy for the time the system needs to recognize actions.

The code for three validation phases is available on GitHub repositories\footnote{\footnotesize{\href{https://github.com/ValerioBelcamino/JAIST_Cylinder}{github.com/ValerioBelcamino/JAIST\_Cylinder}}}
\footnote{\footnotesize{\href{https://github.com/ValerioBelcamino/JAIST_Cylinder_Realtime_Experiment}{github.com/ValerioBelcamino/JAIST\_Cylinder\_Realtime\_Experiment}}} and a video showing one trial of the experiment can be found on our YouTube channel\footnotemark[1].

\section{Results}
Having outlined the scope of our research, the experimental setup and the protocol employed to acquire contact forces and motion data, we now present our results.

\subsection{Offline Validation}
In the first phase of our study, we collected a 4-hour dataset using the TacLINK sensor and TER glove, with 8 participants performing 15 predefined actions for one minute each. The dataset is evenly distributed across all action classes and captures contact forces and hand motions. We employed this data to train and test our model and obtained a final accuracy of 94.64\% with an F1-score of 95.60\%. Additionally, we provide the F1-scores of two unimodal classifiers, namely the IMU-only and video-only models, both trained on the same dataset described above. As reported, the multimodal classifier achieves the highest F1-score outperforming the IMU-only and video-only models, which achieve 91.74\% and 90.66\%, respectively. A more in-depth analysis of these models, along with their detailed descriptions, can be found in our previous work \cite{belcamino2025comparative}.
The model demonstrated exceptional accuracy in classifying actions such as \textit{lingering}, \textit{tapping}, \textit{patting}, \textit{pinching}, \textit{pulling}, \textit{pushing}, \textit{scratching} and \textit{squeezing} achieving accuracies, precisions, and recalls exceeding 97\% for each action. In addition, \textit{poking}, \textit{trembling} and \textit{idle} showed values close to the overall accuracy of 94.1\%, 96.3\% and 95.2\%. On the other hand, for \textit{massaging}, \textit{rubbing}, \textit{shaking}, and \textit{stroking}, our model exhibited accuracies below the global average, achieving scores of 92.11\%, 81.00\%, 90.02\% and 93.31\%, respectively. Additionally, we observed significant crosstalk between \textit{rubbing} and \textit{tapping}, with 9.13\% of samples from the former incorrectly assigned to the latter. Consequently, \textit{rubbing} exhibited the lowest recall (81.56\%), while \textit{tapping} showed the lowest precision (79.86\%). 
This result can be explained by the similarities in hand motion and contact forces that characterize these actions. Specifically, in both cases, the fingers remain almost entirely adducted and slightly bent towards the palm, creating a moderate contact area and pressure from the tips of all four fingers. Furthermore, both actions are performed with a high frequency.

\begin{table}[t]
\centering
\caption{Event-based classification results from online validation. Top: action segments; bottom: idle segments.}
\label{tab:event_based_metrics}
\resizebox{\columnwidth}{!}{\Large
\begin{tabular}{lccccc}
\textbf{Model} & \textbf{True Pos. (\%)} & \multicolumn{4}{c}{\textbf{False Neg. (\%)}} \\
\cmidrule(lr){3-6}
 & & Underfill Start & Underfill End & Fragmentation & Deletion \\
\midrule
IMU-only       & 76.45 & 4.88 & 2.73 & 6.49 & 9.42 \\
Video-only     & 74.12 & 5.20 & 3.10 & 7.02 & 10.56 \\
Multimodal     & 85.97 & 3.04 & 0.99 & 5.58 & 4.42 \\
 & & & &  &  \\
\textbf{Model} & \textbf{True Neg. (\%)} & \multicolumn{4}{c}{\textbf{False Pos. (\%)}} \\
\cmidrule(lr){3-6}
 & & Overfill Start & Overfill End & Insertion & Merge \\
\midrule

IMU-only       & 89.15 & 0.09 & 5.42 & 5.32 & 0.01 \\
Video-only     & 74.37 & 0.15 & 13.78 & 12.60 & 0.10 \\
Multimodal     & 78.96 & 0.07 & 7.87 & 13.10 & 0.00 \\
\end{tabular}
}
\end{table}

\subsection{Online Validation}
In the second phase of our analysis, we used the previously trained model to classify action sequences and compare the output labels with the ground truth. As previously introduced, we started with a frame-by-frame comparison where the model correctly classified 83.92\% frames. Besides the global score, we also computed the accuracy for each action, grouping the ground truth frames by label and computing the percentage separately. In this case, the classes with the highest accuracy were \textit{scratching}, \textit{pinching} and \textit{poking} with 98.31\%, 94.03\% and 93.61\%. 
\textit{Tapping} showed the worst results with 67.82\% accuracy. The low recall (68.82\%) for this class suggests that the model fails to identify a significant portion of its true instances, likely misclassifying them as other actions. This problem arises because \textit{tapping} is frequently mistaken for \textit{patting} due to their similar characteristics.
The model also exhibits performances below the global accuracy for \textit{pushing} and \textit{pulling}, achieving 77.39\% and 76.52\%, respectively. Likely we observe this result because these actions are more affected by the execution time as they are associated with a low frequency, resulting from the hand standing still for most of the action duration.

To further analyze our approach in the online classification, we included the event-based metrics introduced in Section \ref{sec:methodology}. With this criterion, actions are treated as segmented events and associated with a starting and ending time. Depending on the overlap and separations between the ground truth and the classifier's output, we can identify eight different types of errors shown in Table \ref{tab:event_based_metrics}.
The first table illustrates how our approach classifies segments associated with an action label in the ground truth. The classification results are categorized as either true positives or false negatives, with the latter further subdivided into \textit{underfills}, \textit{fragmentation}, and \textit{deletion}. We can observe that the true positive rate (85.97\%) is similar to the accuracy displayed on a frame-by-frame basis. Additionally, most of the false negatives are due to action \textit{fragmentation}, which accounts for 5.58\%, followed by \textit{deletion} at 4.42\%, \textit{underfill\_start} at 3.04\%, and \textit{underfill\_end} at 0.99\%. Instead, the second table reports true negatives (78.96\%) and false positives, with the latter further divided into \textit{insertions} (13.10\%), \textit{overfill\_start} (0.07\%), \textit{overfill\_end} (7.87\%), \textit{merge} (0.00\%). Insertion errors are often related to misclassifications of the \textit{idle} action that, depending on the movement of the hand, could be mistaken as one of the other actions. As for overfills, they all occurred toward the sequence end (\textit{overfill\_end}) and resulted in delays in the sequence classification, most probably induced by the moving window mechanism. Lastly, the lack of mergings is mainly due to the experimental setup, with only a partial contribution from the model's performance. Merges occur between \textit{non-idle} classes, but action repetitions separated by \textit{idle} pauses make merges less likely. Experiments where the person transitions directly from one \textit{non-idle} action to another would better clarify the model's performance concerning this metric.
Fig. \ref{fig:timeplot} shows an example of the event-based metrics from a segment of our continuous dataset. We can notice how the classified blocks are slightly bigger and skewed to the right (i.e., delayed in time) compared to the ground truth. For this reason, the initial errors for each block are generally associated with an \textit{underfill\_start} and the final one with a \textit{overfill\_end}. 
As we did for the previous test, we provide the same classification metrics for the unimodal classifiers. As shown in Table \ref{tab:event_based_metrics}, the multimodal classifier outperforms the video-only model. Although the IMU-only classifier achieves better accuracy for the \textit{idle} action, all other actions are more accurately captured by the combined approach. These findings are further supported by the frame-by-frame accuracies of 83.92\% for the multimodal model, compared to 79.54\% and 71.03\% for the IMU-only and video-only classifiers, respectively.
\begin{figure*}[!t]
\centering
\includegraphics[width=0.75\textwidth]{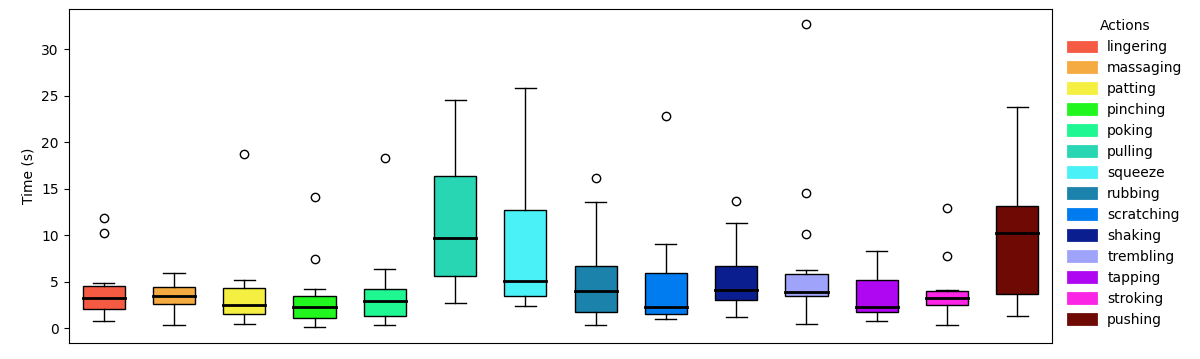}
\caption{The plot refers to the Dynamic Validation of our system. Each box describes the distribution of the classification time for each label. The black line represents the median value, while the black circles represent outliers
}
\label{fig:boxplot}
\end{figure*}

\subsection{Dynamic Validation}
In the dynamic phase, we tested the latency and accuracy of real-time classification within an HRC scenario. For each trial of the experiment, we collected the time needed by the model to detect the correct activity. We should point out that the measured response times start when the experimenter instructs the user on the action to perform and it ends when the robot dynamically responds to the gesture. This design introduced an overhead, as the user first needed to recall the specified action and transition from a resting position, with arms at their sides, to an active stance by approaching the moving sensor. Consequently, these measurements capture both the system's classification latency and the inherent response delay introduced by the experimental setup. Additionally, a small delay is introduced by the moving window approach of the classifier. Based on the enhanced robustness exihibited by the multimodal classifier in the offline and online validations, this final test was carried out using only this model.

Over the 50 performed actions - 5 for each of the 10 subjects - our system took an average of \SI{5.34}{\second} to correctly recognize the action, with a median recognition time of \SI{3.54}{\second} and a standard deviation of \SI{5.44}{\second}. There are no significant differences if we consider trials with left and right hands separately as the average values are \SI{5.87}{\second} and \SI{5.51}{\second} and the standard deviations are \SI{5.07} and \SI{5.39}, respectively. This result suggests that our original training dataset was sufficiently balanced, and the performance of our approach is not affected by the hand which the user intends to use. Similarly, the trajectories performed by the robot do not appear to affect classification times significantly, since the average reaction times are \SI{6.02}{\second} for the square path, \SI{6.46}{\second} for the diamond path, \SI{4.46}{\second} for the hourglass path, \SI{4.42}{\second} for the triangle path, and \SI{5.35}{\second} for the circle path. 
The selected trajectories differ in speed, number of direction changes and working space; however, they still represent a simplification of a real use case, since the motion is limited to a 2D plane. This constraint does not affect the vision-based tactile sensor, whose image reference frame is fixed to the robot’s end effector and thus remains consistent regardless of movement direction. In contrast, IMU features are expressed in the hand’s local frame and can change significantly when the hand assumes different orientations, potentially leading to mismatches with training conditions. Notably, this represents a limitation of our dataset rather than an inherent issue with the classifier itself; support for 3D motion could be achieved through fine-tuning with training data that includes a broader range of orientations. Lastly, we note that translational movement along the Z-axis would not affect performance, as this limitation is primarily linked to orientation changes, not position.
Lastly, we computed the distribution of the classification times for each action independently, and Fig. \ref{fig:boxplot} presents the results. All the actions generally follow the trend of low average and median recognition times, with three notable exceptions: \textit{pulling}, \textit{squeezing}, and \textit{pushing}, which required average recognition times of \SI{10.80}{\second} (SD = \SI{6.47}), \SI{9.63}{\second} (SD = \SI{7.43}), and \SI{10.44}{\second} (SD = \SI{7.75}), respectively. These results are in line with what observed in the online validation. Since the model frequently failed to recognize this action on the first attempt, participants had to act multiple times, increasing the measured reaction time. If we observe the action definitions in Table \ref{tab:action_characteristics}, we can find similarities between these three actions that could explain the classification problem. \textit{Pushing}, \textit{pulling}, and \textit{squeezing} are all associated with a large contact area, strong pressure on the sensing membrane and a low frequency. In particular, the latter is dictated by the long contact time between the hand and the TacLINK and is impaired by the robot's movement, which forces the user to follow its movements while holding the contact. The user is often forced to push or pull the cylinder in different directions to accommodate the robot's motions. Although the training set includes actions performed by interacting with the sensorized link in arbitrary positions, it may not provide sufficient diversity to enable the model to generalize effectively to variations introduced by the robot's motion. Similarly, for the squeezing action, the user is often forced to adjust their grip on the sensor to accommodate its movement, interfering with the classification.

\section{Conclusions}
Our work demonstrates the effectiveness of a multimodal approach for human activity recognition in collaborative robotics, merging tactile and motion data for robust, real-time classification. We tested the approach extensively under multiple conditions, including offline and online validations, followed by a benchmark in a simulated HRC task, where we evaluated its performance under dynamic conditions.
In offline validation, our solution achieved high performance with an F1-score equal to 95.60\% and minimal crosstalk between actions. As for the online validation, the global performance decreased to 83.92\%, which is still satisfactory given the number of classes and the scenario complexities. Furthermore, our analysis with event-based metrics revealed that most errors observed during this phase were mainly classification delays. These delays, introduced by the moving-window approach, are generally less disruptive than missing or misclassifying an action; they typically cause only a slight time shift rather than critical errors, thus maintaining reliable action detection and interpretation.
A possible solution for reducing delays could involve increasing the sensor frequency or adjusting the classifier window length. Future work could focus on optimizing these parameters to enhance the classification pipeline. Additionally, delays should be further investigated through a user study to assess their effect on the user experience. 
Finally, our system was deployed in a dynamic HRC scenario, achieving a median reaction time of \SI{3.54}{\second}. Actions involving prolonged contact, such as pulling, squeezing, and pushing, showed lower classification performance; this could be improved by expanding the training set to cover more dynamic interactions. It is worth noting that the observed delay includes both model latency and user factors, such as reaction time and approach time to the moving robot. While we used a 3-second window to match training sequence length, future versions could reduce this buffer or use overlapping windows to improve responsiveness. As this scenario aimed to evaluate the classifier performance rather than represent a real HRC task, practical deployments would likely involve fewer actions and tighter semantic integration. In such settings, contact onset could be quickly detected using the visual-tactile sensor, supporting responsive safety behaviors while maintaining semantic classification in parallel. Future work should focus on improving classification accuracy and adding the possibility of assigning different functions to each activity depending on whether they occur with or without contact, allowing smoother communication with multiple robots.

\section*{Acknowledgement}
This work was supported by the European Union-NextGenerationEU and by the Ministry of University and Research, National Recovery and Resilience Plan, Mission 4, Component 2, Investment 1.5, project ‘‘RAISE-Robotics and AI for Socio-economic Empowerment’’ (ECS00000035) and partly supported by the Precursory Research for Embryonic Science and Technology PRESTO under Grant JPMJPR2038

\bibliographystyle{IEEEtran}

\bibliography{bibliography}

\end{document}